\documentclass[sigconf]{acmart}
\AtBeginDocument{%
  }

\newcommand{\DUMMY}[1]{#1}

\usepackage{enumitem}
\usepackage{algorithm}
\usepackage{algorithmic}
\usepackage{multirow}
\usepackage[table]{xcolor} % row shading in tables
\newcommand{\field}[1]{\path{#1}}  % breakable, mono-style field names

\usepackage{microtype}
\usepackage{mathtools} % gives \shortintertext
\usepackage{amsmath}
\usepackage{balance}
\setcopyright{acmlicensed}
\copyrightyear{2026}
\acmYear{2026}
\setcopyright{cc}
\setcctype{by}
\acmDOI{10.1145/3770855.3817491}

\acmConference[KDD 2026] {the 32nd ACM SIGKDD Conference on Knowledge Discovery and Data Mining V.2}{August 9--13, 2026}{Jeju Island, Republic of Korea.}

\acmISBN{978-1-4503-15322-80/2026/08}

\begin{document}

\title{From Statute to Control Flow: Span-Grounded Deontic Trees for Defeasible Scope Parsing}

\settopmatter{authorsperrow=4}
\author{Jian Chen}
\authornote{Both authors contributed equally to this research.}
\orcid{1234-5678-9012}
\affiliation{%
  \institution{The Hong Kong University of Science and Technology (Guangzhou)}
  \city{Guangzhou}
  \country{China}
}
\email{jchen524@connect.hkust-gz.edu.cn}

\author{Siyuan Li}
\authornotemark[1]
\affiliation{%
  \institution{The Hong Kong University of Science and Technology (Guangzhou)}
  \city{Guangzhou}
  \country{China}
}
\email{sli974@connect.hkust-gz.edu.cn}

\author{Chucheng Wan}
\affiliation{%
  \institution{Sun Yat-Sen University}
  \city{Guangzhou}
  \country{China}
}
\email{chucheng.wan@outlook.com}

\author{Zixuan Yuan}
\authornote{Corresponding author.}
\affiliation{%
  \institution{The Hong Kong University of Science and Technology (Guangzhou)}
  \city{Guangzhou}
  \country{China}
  }
\email{zixuanyuan@hkust-gz.edu.cn}

\renewcommand{\shortauthors}{Jian Chen, Siyuan Li, Chucheng Wan, and Zixuan Yuan}

\begin{abstract}
Rule-following agents tasked with executing policies and regulations often fail via \emph{Silent Scope Omission (SSO)}: a model applies a general rule but silently drops nested exceptions or counter-exceptions, producing outputs that appear compliant yet break on important edge cases. Although such failures are often framed as an agentic-systems problem, the underlying bottleneck is statutory and policy understanding---a capability typically studied in legal NLP. However, most existing legal NLP benchmarks emphasize end-task outcomes, which can overlook the structural omissions that cause SSO.
To diagnose and mitigate SSO, we introduce \textbf{NormBench} \footnote{https://github.com/AlexJJJChen/NormBench}, a benchmark of 2{,}290 provisions spanning Chinese (laws and local policies), English (U.S.\ tax law, GDPR, and corporate policies), and cross-lingual settings, designed for defeasible scope parsing---identifying precisely which clause overrides which. \textbf{NormBench} uses \textbf{Span-Grounded Deontic Trees (SG-DT)}, a compiler-style intermediate representation that anchors every logical branch to source spans and requires explicit exclusion guards, enabling deterministic compilation and audit.
Evaluations of frontier LLMs reveal two recurring pathologies: (1) \emph{Recursion Decay}, where performance drops sharply as defeater depth increases, and (2) an \emph{Auditability Trap}, where models retrieve relevant spans but fail to assemble correct control flow. Using SG-DT as a constrained intermediate output improves whole-tree fidelity and defeater recovery, and downstream experiments show that its utility is mechanism-specific: gains concentrate on exception-active, SSO-prone cases, while aggregate accuracy can be mixed when the added structure is unnecessary or parser fidelity is low.
\end{abstract}




% \received{08 February 2026}
% \received[accepted]{16 May 2026}
\maketitle

\section{Introduction}
\label{sec:intro}

\begin{figure*}
    \centering
    \includegraphics[width=\linewidth]{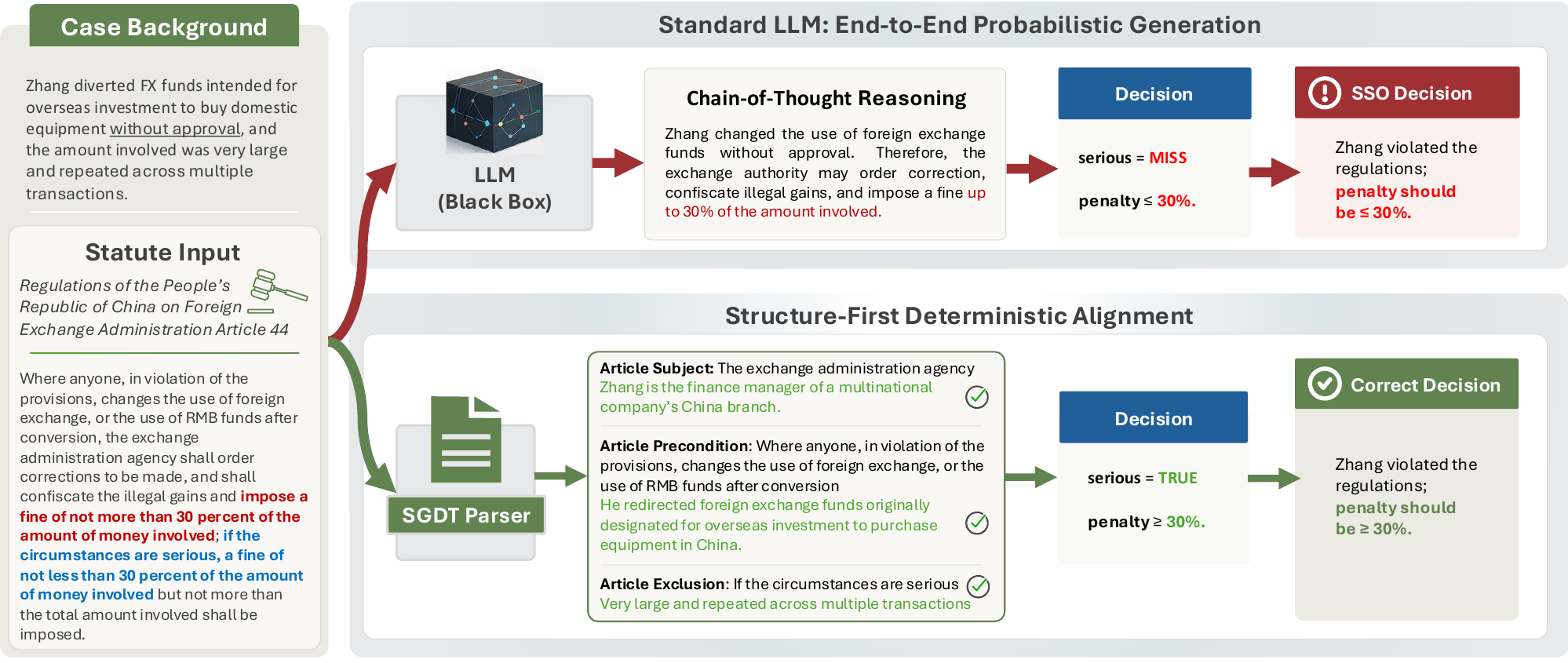}
    \caption{\textbf{Non-monotonic rules and silent omissions.}
    In statutes and policies, later clauses can narrow or override earlier ones (e.g., “unless…”, “except…”, “notwithstanding…”).
    Standard LLM prompting often produces a plausible decision while skipping a nested exception, which we refer to as \textbf{Silent Scope Omission (SSO)}.
    % \textbf{NormBench} requires a \textbf{Span-Grounded Deontic Tree (SG-DT)} as an intermediate output so that each branch is explicitly tied to a source span before it is compiled or executed.
    }
    \label{fig:teaser}
\end{figure*}

Rule-following agents are increasingly used to execute policies and regulations.
Examples include tax assistants that compute penalties, privacy agents that process deletion requests, and enterprise bots that enforce access or reimbursement rules \cite{li2025compliance, freiberger2025you}.
Large Language Models (LLMs) make these systems easier to build, but they also introduce a reliability issue \cite{park2025critical}.
A model can write fluent legal text or even produce syntactically correct code, yet still miss the constraints that actually determine compliance \cite{dahl2024large, zhang2025towards}.
This is especially problematic when the output looks reasonable and therefore passes light-weight checks.

A key reason is that many legal rules are \emph{non-monotonic} \cite{prakken1996dialectical}.
Later clauses can override earlier conclusions, and exceptions can themselves have counter-exceptions \cite{gordon1995pleadings}.
In practice, we often see the following failure:
the model applies a general rule but does not materialize a nested exception (or counter-exception).
We call this \textbf{Silent Scope Omission (SSO)} \cite{kim2025does}.
SSO tends to hide in edge cases, where correctness depends on reading and applying the tail-end “defeater” clauses (Figure~\ref{fig:teaser}).

Rule-following agents ultimately need \emph{checkable} reasoning, because their outputs are acted upon \cite{mohun2020cracking}.
What matters is not only a final decision, but whether the system can expose which clause is active, which exception applies, and how conflicts are resolved within a provision.
This links rule-following agents to legal NLP: both rely on statutory understanding, especially recovering how later clauses restrict earlier ones.
However, most legal NLP benchmarks emphasize \emph{end-task outcomes} such as QA, entailment, or judgment prediction \cite{guha2023legalbench, yu2025benchmarking, fan2025lexam}.
These settings can reward a correct final answer even when the model fails to recover the provision’s internal scope/override structure.
For agents, that hidden structure is exactly what determines whether an exception was properly handled.
We therefore need a \emph{diagnostic} benchmark that directly evaluates clause-level scope and override mechanics, complementing downstream legal reasoning tasks.

To address this gap, we introduce \textbf{NormBench}, a benchmark for normative scope parsing and compilation.
NormBench contains 2,290 provision-language items across multiple languages and domains.
Each item is annotated with a \textbf{Span-Grounded Deontic Tree (SG-DT)}, a structured intermediate representation that makes exceptions and override relations explicit.
SG-DT grounds every node and branch to an immutable text span.
This design supports audit (the provenance of each decision is visible) and also supports compilation into executable decision procedures for compliance-to-code systems \cite{li2025compliance}.

NormBench is organized from \emph{controlled} to \emph{in-the-wild}.
It starts from codified statutory provisions in Chinese (NormBench-ZH), includes official ZH--EN translation pairs, and adds an English slice (NormBench-EN) covering U.S. tax law, the EU GDPR, and corporate policies.
This layout lets us test generalization while controlling for language and jurisdictional changes.
When clause overlap or precedence is not determined by the text alone, we mark this explicitly as metadata rather than forcing an arbitrary resolution.
This allows evaluation to separate true scope understanding from cases that are genuinely under-specified.

We use \textsc{NormBench} to perform a deep pathology analysis of frontier LLMs (e.g., DeepSeek-V3.2 \cite{liu2025deepseek}, GPT-5.2 \cite{OpenAI_2025}, Claude-Opus-4.5 \cite{Anthropic_2025}). Our experiments yield three critical insights:
\begin{enumerate}[leftmargin=*, labelsep=0.4em]
  \item \textbf{The Recursion Bottleneck:} Performance degradation is driven by logical depth, not context length. Even when controlling for token count, models struggle significantly more with nested exceptions ($D \ge 2$) than with linear rules, suggesting a reasoning limitation distinct from retrieval capacity.
  \item \textbf{The Structure-Grounding Gap:} While models are proficient at identifying relevant text spans, they struggle to attach them to the correct logical parent in the deontic tree. This confirms that ``finding the rule'' is easier than ``understanding its scope.''
  \item \textbf{Mechanism-specific downstream utility:} SG-DT is not a drop-in scaffold for all legal reasoning cases. Its gains concentrate on exception-active / SSO-prone examples where the outcome depends on applying a defeater; on simpler cases, additional structure can impose overhead.
\end{enumerate}

\noindent\textbf{In summary, our contributions are:}
\begin{itemize}[leftmargin=*, labelsep=0.4em]
  \item \textbf{\textsc{NormBench} \& SG-DT.} A standardized benchmark and intermediate representation for evaluating structural scope recovery in legal texts, covering 2{,}290 items across controlled and in-the-wild domains.
  \item \textbf{Pathology analysis.} We identify the Recursion Bottleneck and Auditability Trap as key barriers to reliable legal agents, separating logical failures from retrieval failures.
  \item \textbf{Mechanism-specific intervention.} We show that SG-DT is most useful when outcomes depend on active defeaters, while aggregate downstream gains can be mixed when parser fidelity is low or added structure is unnecessary.
\end{itemize}

\section{Related Work}
\label{sec:related}

\paragraph{Statutory reasoning benchmarks and the missing \emph{scope} diagnosis.}
A recurring deployment risk in statutory reasoning is not lack of fluency but \emph{silent scope failure}: models issue confident decisions while mis-attaching exceptions, mis-ordering overrides, or collapsing nested defeaters inside a single provision \cite{yu2025benchmarking}.
Classic benchmarks operationalize the problem as end-to-end decision making---e.g., conversational rule interpretation or follow-up QA (ShARC, SARA) \cite{saeidi2018interpretation, holzenberger2020dataset}---and broad suites expand coverage across legal reasoning types (LexGLUE, LegalBench) \cite{chalkidis2022lexglue, guha2023legalbench}.
However, answer-level supervision rarely isolates whether the model recovered the \emph{major premise} correctly, so scope/defeat errors can be rewarded when they coincidentally lead to the right label. \cite{ariai2025natural}
Prior analyses show that exposing intermediate statutory structure improves outcomes \cite{holzenberger2021factoring}, motivating NormBench to make span-grounded intra-provision scope/defeat (exceptions-to-exceptions, precedence/override) an explicit prediction and evaluation target.

\paragraph{RAG-style legal agents and comprehensive evaluation.}
Retrieval-Augmented Generation (RAG) has been widely used in various NLP tasks \cite{chen2024fintextqa, chen2025dekeynlu}. Modern legal assistants are increasingly implemented as RAG pipelines, where retrieval determines which clauses (often decisive exceptions) are even visible to the generator \cite{kim2025does}.
This makes silent-scope failures harder to diagnose: an incorrect final answer can arise either because retrieval missed the controlling exception, or because the model saw it but silently dropped or mis-scoped it \cite{li2024lexeval}.
Large-scale suites such as LexEval \cite{li2024lexeval} and LegalBench \cite{guha2023legalbench} provide broad downstream coverage, but typical aggregate metrics still obscure this attribution.
We therefore use them as an external stress test while explicitly decomposing oracle vs.\ real retrieval and reporting a silent-scope risk metric; NormBench complements downstream evaluation by directly supervising normative structure recovery.

\paragraph{Normative formalisms, law-as-code, and span-faithful parsing.}
Compliance-critical systems need \emph{executability} \emph{and} \emph{auditability}: every branch of a decision procedure should be traceable to the precise statutory spans that authorize it \cite{athan2015legalruleml}.
Normative standards and DSLs (e.g., LegalRuleML, Catala) offer rigor and explainability \cite{athan2015legalruleml, huttner2022catala}, but typically require expert authoring and often abstract away from token-level provenance, limiting data-scale learning from raw statutes.
Conversely, approaches that directly generate executable code from text \cite{li2025compliance} can widen an auditability gap: control flow may be difficult to map back to the exact exception/override spans, allowing scope omissions to pass superficial checks.
NormBench positions Span-Grounded Deontic Trees (SG-DT) as an agent-facing intermediate representation that preserves immutable span pointers while making recursive defeasibility and attachment explicit, going beyond surface component extraction (RASE-style) \cite{hjelseth2011capturing}, non-deontic meaning representations (AMR) \cite{banarescu2013abstract}, and document-structure markup (Akoma Ntoso) \cite{vitali2007towards}.

\section{Span-Grounded Deontic Trees (SG-DT)}
\label{sec:schema}

% SG-DT is an \emph{agent-facing intermediate representation (IR)} for statutory provisions that is simultaneously
% (i) \emph{span-faithful} (every node is an immutable pointer to the verbatim source text) and
% (ii) \emph{compilable} into deterministic control flow for downstream executors.

\subsection{Design Goals and Scope}
\label{sec:schema:goals}
SG-DT is engineered around four constraints that standard free-form generation routinely violates:
\textbf{(G1) Span-grounded auditability \& executability:} each logical node is grounded in the statute and can be deterministically mapped to downstream signatures (checks / code blocks).
\textbf{(G2) Decidability under defeasibility:} ``unless/except/provided that'' are compiled into explicit guards so execution never depends on implicit precedence that can silently drop tail-end scope.
\textbf{(G3) Deterministic evaluation:} the same structure supports exact structural scoring and deterministic compilation under a single schema.
\textbf{(G4) Learnability:} the representation is dense and seq2seq-friendly, avoiding expert-only ontologies.

When a provision contains multiple separable normative points, we optionally split it into a few contiguous verbatim segments (``executable units'') as the alignment scope; short single-point provisions are treated as a single segment.

\paragraph{Non-goals.}
SG-DT is not a full legal KR language: we do not encode open-textured standards, teleological reasoning, or domain ontologies beyond what is necessary for dataset-scale, auditable compliance checking.
Importantly, \emph{mutual exclusivity} in SG-DT is a \emph{compilation discipline} (for execution/evaluation), not a claim that real legal exceptions are inherently exclusive.
When provisions contain overlap, SG-DT determinizes via explicit \emph{case partitions} and records optional span-grounded justifications for any refinement/adjudication choice.

\paragraph{Terminology.}
We use \emph{defeater} as the general term for any span that narrows, overrides, or blocks the application of another normative branch.
An \emph{exception} is a defeater that blocks a general rule and is represented as a positive \textsc{Precondition} in the exception branch plus an \textsc{Exclusion} guard in the general branch.
A \emph{proviso} is surface drafting language such as ``unless'', ``except'', or ``provided that''; in SG-DT it is normalized into the same branch/exclusion structure.
A \emph{counter-exception} is a defeater attached to another defeater, producing a deeper defeat chain.
We therefore distinguish surface cues from their SG-DT roles: surface provisos trigger typed leaves, while executable disjointness is represented by \textsc{Exclusion} guards.

\subsection{Schema}
\label{sec:schema:definition}
Let $X=\{x_1,\dots,x_n\}$ be the token sequence of the current text scope (a full provision, or an executable unit when segmentation is applied).
An SG-DT annotation is $\mathcal{S}=(X,\mathcal{B})$ where $\mathcal{B}=\{B_i\}_{i=1}^m$ is a set of \textit{Normative Branches}.
Each branch is
\begin{equation}
B_i=(\alpha,\kappa,\mathcal{C},\mathcal{E}),
\end{equation}
optionally carrying an \emph{audit payload} $\mathcal{J}(B_i)$, which does not change the logical content of $B_i$.

\begin{itemize}[leftmargin=*, labelsep=0.4em]
  \item \textbf{Anchor ($\alpha$):} a span $X[i\!:\!j]$ locating the normative trigger (e.g., ``shall pay''), disambiguating repeated phrases.
  \item \textbf{Modality ($\kappa$):} force drawn from a typed set $\mathcal{K}$ (e.g., \textsc{Obligation}, \textsc{Prohibition}, \textsc{Permission}, \textsc{Right}, \textsc{Liability}, \textsc{Sanction}, \textsc{Definition}, \textsc{Procedure}, \textsc{Other}).
  \item \textbf{Condition tree ($\mathcal{C}$):} boolean composition over typed, span-grounded leaves:
  \begin{equation}
  \mathcal{C} ::= \ell \mid (\mathcal{C}\land\mathcal{C}) \mid (\mathcal{C}\lor\mathcal{C}),
  \qquad \ell=(\text{type},\text{span}).
  \end{equation}
  \item \textbf{Effects ($\mathcal{E}$):} a set of span pointers representing the normative consequence when $\mathcal{C}$ holds.
\end{itemize}

\paragraph{Serialization.}
While the definition uses token spans $X[i\!:\!j]$, the released format serializes spans as deterministic \emph{text pointers} within an executable unit (e.g., \texttt{(text, occurrence)}) to disambiguate repeated strings.
Leaves are \texttt{(tag, text)} with \texttt{tag} drawn from a fixed schema-level inventory (e.g., \textsc{Subject}, \textsc{Action}, \textsc{Precondition}, \textsc{Time}, \textsc{Threshold}, \textsc{Reference}, \textsc{Exclusion}, \dots).
\textsc{Exclusion} is reserved exclusively for compilation-introduced disjointness (Section~\ref{sec:schema:defeasibility}).
The optional payload $\mathcal{J}$ stores a small set of justification pointers.

\paragraph{Typed leaves as atomic predicates.}
Although $\mathcal{C}$ exposes only $\land/\lor$, each leaf type induces a downstream predicate
$p_{\text{type}}(\text{span},\text{context})\rightarrow\{0,1\}$ (e.g., roles, thresholds, time limits, references).
This separates decidable boolean structure from domain-dependent interpretation, and keeps nested provisos/exception chains representable via \emph{branch decomposition} rather than higher-order operators.

% \subsection{Construction: Unitize, Then Compile}
% \label{sec:schema:process}
% Direct parsing of long provisions is brittle. SG-DT mirrors legal analysis in two deterministic stages.

% \paragraph{Stage 1: Executable unitization.}
% We split a provision into \emph{Executable Units} (contiguous segments that can be unit-tested), reducing long-range coreference and ambiguity.
% We release deterministic unitization guidelines (decision table + pseudocode) covering enumerations, provisos, embedded definitions, and related cues.
% Stability is tested with \emph{coarse} vs.\ \emph{fine} strategies: depth statistics and model rankings remain stable up to $\pm2$ Tree-ES points and $\pm0.02$ SSDR (Appendix~\ref{app:unit_stability}).

% \paragraph{Stage 2: Branch compilation.}
% Each unit is compiled into one or more branches, making implicit scope explicit (general rule vs.\ provisos/exceptions).
% Compilation is deterministic: branches are ordered by anchor occurrence, and whenever a unit contains defeaters, we enforce disjoint control flow using explicit exclusion guards (Section~\ref{sec:schema:defeasibility}).

\subsection{Defeasibility via Explicit Exclusion}
\label{sec:schema:defeasibility}
Statutory rules are \emph{defeasible}: a rule applies unless an exception applies. Free-form parsers often treat exceptions as optional modifiers, enabling silent scope omission in downstream code.
SG-DT prevents this with an \textit{Explicit Exclusion Policy}. For a rule with exception trigger $T$:
\begin{enumerate}[leftmargin=*, labelsep=0.4em]
  \item \textbf{General branch:} includes an \textsc{Exclusion} leaf pointing to $T$, restricting scope to $\neg T$.
  \item \textbf{Exception branch:} includes $T$ as a positive \textsc{Precondition}.
\end{enumerate}
Thus $B_{\text{gen}}\land B_{\text{exc}}\equiv \text{False}$ by construction, without heuristic priority rules.
We define compiled depth $D$ as the longest exclusion-induced defeater chain: $D=0$ for no exception, $D=1$ for a direct exception, and $D\ge2$ for nested defeaters or counter-exceptions.

\paragraph{Multiple and overlapping exceptions.}
Real provisions may include triggers $\{T_1,\dots,T_k\}$ that partially overlap. SG-DT does not assume legal exclusivity; it determinizes only as needed for execution using \emph{minimal disjoint refinement}:
\begin{alignat}{3}
& B_0         \quad &&:\quad \neg(T_1 \lor \cdots \lor T_k) && \\
& B_i         \quad &&:\quad T_i \land \neg\bigvee_{j<i} T_j
  && \parbox[t]{.44\columnwidth}{\raggedright\footnotesize ordered partition, \emph{only with explicit priority cue}}\\
& B_{i\wedge j}\quad &&:\quad (T_i \land T_j)
  && \parbox[t]{.44\columnwidth}{\raggedright\footnotesize intersection branch, \emph{only if overlap has distinct consequence}}
\end{alignat}
Outcome-equivalent overlap may be OR-merged; distinct-outcome overlap introduces an explicit intersection branch.
Precedence-based ordering is permitted \emph{only} with textual evidence; any such refinement attaches span-grounded justification $\mathcal{J}$.

\paragraph{Gold protocol under overlap (G1--G4, summary).}
(i) outcome-equivalent overlap $\Rightarrow$ OR-merge;
(ii) distinct-outcome overlap $\Rightarrow$ explicit intersection;
(iii) precedence only with an explicit textual priority cue (and justification spans);
(iv) otherwise surface uncertainty via an \texttt{Ambiguous} flag and adjudication notes (optionally with alternative refinements).

\subsection{Declarative vs.\ Executable Semantics}
\label{sec:schema:semantics}
Statutes can be declaratively co-applicable, while executors require deterministic control flow.
SG-DT separates these views and makes all determinization decisions inspectable.

\paragraph{Declarative meaning.}
For a unit $u$ with branches $\mathcal{B}_u$, each branch denotes the span-grounded pair $(\mathcal{C}_i,\mathcal{E}_i)$: if predicates induced by $\mathcal{C}_i$ hold under a downstream interpretation, effects $\mathcal{E}_i$ trigger. Declaratively, multiple branches may hold simultaneously.

\paragraph{Executable semantics.}
Executable semantics is a deterministic control-flow program whose guards are induced by $\mathcal{C}$ and made disjoint by compilation (\textsc{Exclusion} leaves + minimal refinement), avoiding hidden priority rules.

\paragraph{What SG-DT guarantees.}
SG-DT guarantees \emph{no silent omission at the IR level}: every defeater trigger span must appear as an explicit node and must induce either a positive exception guard or a negated exclusion guard; missing triggers become validator-detectable structural errors.
SG-DT does \emph{not} guarantee unique executability when determinization is inherently extra-textual (e.g., open-textured standards, concurrent carve-outs without ordering cues, multi-authority conflicts). In such cases it surfaces uncertainty (\texttt{Ambiguous}/alternatives) rather than hard-coding a choice.

\subsection{Comparison with Related Standards}
\label{sec:schema:comparison}
SG-DT does not replace full KR/DSL standards; it is optimized for \textit{span-level supervision and auditing}.
As shown in Table \ref{tab:comparison}, LegalRuleML and Catala provide richer execution semantics but typically require manual predicate engineering and detach from source spans; Akoma Ntoso models document structure/metadata rather than executable normative logic.
SG-DT complements these by providing lossless span grounding, explicit defeasible scope, and auditable, deterministic compilation refinements.

\begin{table}[t]
\centering
\small
\caption{\textbf{Representation trade-offs.} SG-DT targets span-grounded defeasible scope for supervised learning and auditing.}
\label{tab:comparison}
\resizebox{\columnwidth}{!}{
\begin{tabular}{l|cccc}
\toprule
\textbf{Feature} & \textbf{LegalRuleML} & \textbf{Catala (DSL)} & \textbf{Akoma Ntoso} & \textbf{SG-DT (Ours)} \\
\midrule
Primary goal & KR/Execution & Execution & Doc structure & IR + Auditing \\
Span grounding & No & No & Partial & \textbf{Yes} \\
Defeasible scope & Yes & Yes & No & \textbf{Yes} \\
Deterministic compilation & Yes & Yes & N/A & \textbf{Yes} \\
\bottomrule
\end{tabular}
}
\end{table}

\subsection{Downstream Utility}
\label{sec:schema:downstream}
\paragraph{Traceable reasoning.}
$\mathcal{C}$ isolates factual prerequisites; reasoning reduces to verifying typed leaf predicates against case facts, with provenance via the satisfied path in $\mathcal{C}$.

\paragraph{Constrained code generation.}
SG-DT provides an executable specification for synthesis: modality $\kappa$ determines function signatures, and disjoint branches induce control flow. Supervising SG-DT before code narrows the search space and reduces silent omission of tail-end exceptions.

\paragraph{One schema, two consistent views.}
For benchmarking, SG-DT is a span-grounded structural target that makes failures measurable (missing exclusion guards, mis-attached provisos, unjustified precedence).
For compilation, the same structure yields deterministic control flow without heuristic priority reasoning; any refinements are recorded via $\mathcal{J}$.

\section{The NormBench Dataset}
\label{sec:dataset}

NormBench benchmarks \emph{intra-provision} defeasible scope parsing in rule texts.
Given a provision $X$, systems output a Span-Grounded Deontic Tree (SG-DT) $S=(X,\mathcal{B})$ with nodes as pointers to spans in $X$ (Sec.~\ref{sec:schema}).
The span-grounded structure is auditable and deterministically compilable, directly targeting Silent Scope Omission.

\subsection{Data Collection and Source Coverage}
\label{sec:dataset:slices}

\noindent\textbf{Design principle.}
All slices share the \emph{same} SG-DT schema, validator, and metrics, so differences in performance can be attributed to drafting and domain shift rather than a moving evaluation contract.
% (Counts and composition are reported in Table~\ref{tab:dataset_stats}.)

\paragraph{NormBench-ZH (codified statutory core).}
We collect Chinese provisions from the China Laws \& Regulations Database (CLRD)\footnote{https://flk.npc.gov.cn/index}, focusing on codified legal and policy documents, to construct a controlled distribution in which provisions are relatively self-contained and scope cues are explicit.
Sampling is stratified across hierarchy strata (e.g., national laws, administrative regulations, and local policies) to introduce drafting variability within a single jurisdictional baseline.
We prioritize rule-like provisions that contain at least one explicit defeater cue (e.g., ``unless'', ``except'', ``notwithstanding'', ``provided that''-style provisos) and exhibit non-trivial defeasible interaction (nested exceptions, counter-exceptions, or overlap/precedence patterns).

\paragraph{NormBench-EN (in-the-wild English rules).}
To stress generalization beyond codified drafting, we collect English rule texts from three domains with distinct styles: 
U.S.\ tax law \footnote{https://www.irs.gov}, 
EU GDPR \footnote{https://gdpr.eu}, 
and corporate policies \footnote{https://www.apple.com/legal/}.
We focus on provision-/article-level excerpts that are reasonably self-contained, contain explicit exception/proviso language, and yield executable case partitions after SG-DT compilation.
This slice deliberately introduces domain shift (specialized definitions, cross-article references, and operational policy language) while keeping the annotation contract identical.

\paragraph{Official ZH--EN parallels (cross-lingual invariance).}
For a subset of the ZH items, we pair the original Chinese provision with its authoritative English translation from the Ministry of Justice of the People's Republic of China \footnote{http://en.moj.gov.cn/} to test whether systems recover the same underlying scope structure across languages.

\begin{table}[t]
    \centering
    \small
    \caption{\textbf{NormBench dataset statistics (extended).} Supplements key structural phenomena with unit/branch totals.}
    \label{tab:dataset_stats}
    \resizebox{\columnwidth}{!}{
    \begin{tabular}{l|r|l}
    \toprule
    \textbf{Dimension} & \textbf{Count / \%} & \textbf{Notes} \\
    \midrule
    \multicolumn{3}{l}{\textit{\textbf{A. Composition}}} \\
Total items & \textbf{2,290} & shared SG-DT contract \\
Unique underlying provisions & 1,790 & EN parallels are translations \\
NormBench-ZH & 1,524 & P.R.C. codified provisions \\
Official ZH--EN parallels & 500 pairs & authoritative EN translations \\
NormBench-EN & 266 & us\_taxlaw=69 / gdpr=95 / company\_terms=102 \\
    \midrule
    \multicolumn{3}{l}{\textit{\textbf{B. ZH source types}}} \\
Law & 1,248 (82\%) & national statutes (NPC/SC) \\
Local Policy & 163 (11\%) & by Local People’s Congress \\
Judicial Interpretation & 113 (7\%) & SPC/SPP judicial interpretations \\
    \midrule
    \multicolumn{3}{l}{\textit{\textbf{C. Core complexity (All)}}} \\
Level 1 (Simple) & 834 (36\%) & mostly linear \\
Level 2 (Nested) & 610 (27\%) & one defeater layer \\
Level 3+ (Recursive) & \textbf{846 (37\%)} & deep/counter-exceptions \\
    \midrule
    \multicolumn{3}{l}{\textit{\textbf{D. Key structural phenomena (All, span-grounded)}}} \\
Avg. branches / unit & 1.63 & compiled branches per unit \\
Median branches / unit & 1 &  \\
Units w/ explicit exclusion & 9\% & triggers Exclusion policy \\
Units w/ cross-article refs & 3\% & scope depends on references \\
Total units & 5,543 & sum over all units \\
Avg units / article & 2.42 & total units / number of articles \\
Total branches & 9,019 & sum over all unit branches \\
Avg branches / article & 3.94 & total branches / number of articles \\
    \bottomrule
    \end{tabular}}
\end{table}

\subsection{SG-DT-aligned decomposition}
\label{sec:dataset:alignment}

% \noindent\textbf{Two-stage alignment to make defeasibility decidable.}
% \label{sec:schema:process}
Direct parsing of long provisions is brittle. SG-DT mirrors legal analysis in two deterministic stages.

\paragraph{Stage 1: Executable unitization.}
We split a provision into \emph{Executable Units} (contiguous segments that can be unit-tested), reducing long-range coreference and ambiguity.
We release deterministic unitization guidelines (decision table + pseudocode) covering enumerations, provisos, embedded definitions, and related cues.
Stability is tested with \emph{coarse} vs.\ \emph{fine} strategies: depth statistics and model rankings remain stable.

\paragraph{Stage 2: Branch compilation.}
Each unit is compiled into one or more branches, making implicit scope explicit (general rule vs.\ provisos/exceptions).
Compilation is deterministic: branches are ordered by anchor occurrence, and whenever a unit contains defeaters, we enforce disjoint control flow using explicit exclusion guards (Section~\ref{sec:schema:defeasibility}).

\paragraph{Derived defeater-attachment projection.}
\label{sec:dataset:transfer_skeleton}
All NormBench items are annotated with the \emph{full} SG-DT contract (typed leaves + explicit exclusion).
To isolate defeater attachment and recursion under drafting shift, we additionally release a deterministic projection computed from gold SG-DTs that retains:
(i) unit boundaries, (ii) exception trigger spans $\{T_i\}$, and (iii) directed attachment/defeat relations (including exception-to-exception).
Because the view is computed from gold, it introduces no new annotation noise while directly testing recovery of the \emph{scope graph} that drives SSO.

\paragraph{Difficulty.}
Beyond recursion depth, we provide a reproducible three-level difficulty label for each \emph{unit}, computed only from released structural signals:
branch count $B$, text length $L$, condition-tree leaf count $C$, max depth $d_{\text{cond}}$, \texttt{has\_or}, and indicators for exceptions, cross-article dependencies, unresolved references, multiple effects, and multiple norm kinds.
We assign an additive score $S$ and map to difficulty via default thresholds:
$S\le 2$ (Low), $3\le S\le 5$ (Medium), and $S\ge 6$ (High).
This yields a stable hardness measure independent of any single model's error profile.

\subsection{Dataset annotation}
\label{sec:dataset:qa}

\noindent\textbf{Proposer--Refiner with expert adjudication.}
We use a model-in-the-loop \emph{Proposer--Refiner} workflow: a strong LLM drafts candidate SG-DTs for triage, and six legally trained annotators (Master/J.D.\ candidates) audit, correct, and adjudicate every released sample. In aggregate, producing 2,290 full SG-DTs required approximately \textbf{\DUMMY{232}} expert-hours (including double annotation and adjudication), with a heavy tail driven by the recursion frontier ($D\ge 4$).
A stratified audit finds that only \textit{40\%} of proposer drafts are structurally correct without edits ($80/200$; 95\% Wilson CI $[0.33,0.47]$), so \textit{60\%} require expert correction.
To mitigate model-induced selection bias, we additionally run a human-only from-scratch control slice and an alternate-proposer sensitivity analysis, summarized in Section~\ref{sec:dataset:quality_controls}.

\noindent\textbf{Ambiguity annotation.}
When overlap/precedence or reference closure is underdetermined by spans, we label units as \texttt{Ambiguous}.
Evaluation is ambiguity-aware: systems may explicitly flag ambiguity, and we provide multi-gold alternatives for key overlap/precedence cases to avoid rewarding brittle, ungrounded commitments.

\noindent\textbf{Reliability.}
Complex units undergo blind double annotation followed by adjudication.
Because SG-DT requires simultaneously correct span pointers \emph{and} global defeater attachment, initial IAA is expected to be moderate for deeper/ambiguous units.
We therefore report both initial agreement (true IAA) and post-reconcile agreement (gold convergence after expert adjudication) under a canonicalized span-grounded TreeMatch definition. The result of IAA is shown in Table \ref{tab:iaa}.

\begin{table}[t]
\centering
\scriptsize
\setlength{\tabcolsep}{4pt}
\caption{\textbf{Annotation reliability.} Initial = pre-reconciliation IAA; Post = gold convergence after adjudication.}
\label{tab:iaa}
\begin{tabular}{p{0.60\columnwidth}ccc}
\toprule
\textbf{Metric} & \textbf{Initial} & \textbf{Post} & \textbf{Target} \\
\midrule
Deontic Label $\kappa$ & 0.76 & 0.91 & $>0.80$ \\
Span Exact Match & 81.2\% & 92.4\% & $>90\%$ \\
Skeleton EM (attach.) & 65.9\% & 86.8\% & $>85\%$ \\
Tree Structure Match & 70.8\% & 88.6\% & $>85\%$ \\
\bottomrule
\end{tabular}
\vspace{-0.8em}

\end{table}

% \subsection{Difficulty stratification and statistics}
% \label{sec:dataset:difficulty}

% Beyond recursion depth, we provide a reproducible three-level difficulty label for each \emph{unit}, computed only from released structural signals:
% branch count $B$, text length $L$, condition-tree leaf count $C$, max depth $d_{\text{cond}}$, \texttt{has\_or}, and indicators for exceptions, cross-article dependencies, unresolved references, multiple effects, and multiple norm kinds.
% We assign an additive score $S$ and map to difficulty via default thresholds:
% $S\le 2$ (Low), $3\le S\le 5$ (Medium), and $S\ge 6$ (High).
% This yields a stable, representation-driven hardness measure independent of any single model's error profile.

\subsection{Dataset Statistics and Characteristics}
\label{sec:dataset:stats}

As shown in Table \ref{tab:dataset_stats}, NormBench provides full SG-DT supervision for 2,290 provision-language items spanning a controlled ZH statutory core (1,524), an official ZH--EN parallel slice (500 pairs), and an out-of-domain EN slice (266).
This corpus covers 1,790 unique underlying provisions and yields a total of 9,019 compiled normative branches (averaging 3.94 branches per article).
Beyond scale, Table~\ref{tab:dataset_stats} highlights the structural complexity inherent in \emph{defeasible scope parsing}. 
A significant portion of the dataset is \emph{recursive} (Level 3+, 37\%), where correctness depends on resolving deep exception-to-exception attachments rather than isolated span recognition.
Additionally, precise control-flow partitioning is essential to handle explicit exclusion guards (9\%) and non-local cross-article references (3\%), ensuring that systems materialize disjoint branches rather than relying on shallow paraphrasing.

% \subsection{Dataset Statistics and Characteristics}
% \label{sec:dataset:stats}

% NormBench provides full SG-DT supervision for \textbf{2,300} provision-language items spanning a controlled ZH statutory core, an official ZH--EN parallel slice, and an out-of-domain EN slice, yielding approximately \textbf{\DUMMY{3{,}250}} compiled normative branches.
% Beyond size, Table~\ref{tab:dataset_stats} highlights two properties that matter for 
% \emph{defeasible scope parsing}: a heavy tail in recursion and a non-trivial rate of scope constructs that force explicit control-flow partitioning.
% In the ZH core, a substantial fraction of units are \emph{recursive} (Level 3+, \textbf{25\%}), meaning correctness depends on resolving exception-to-exception attachment rather than recognizing isolated exception spans.
% Moreover, many units require explicit exclusion guards (\textbf{\DUMMY{33\%}}), which makes ``reasonable'' paraphrases insufficient: systems must materialize disjoint branches to avoid silent omissions.
% Finally, cross-article references (\textbf{\DUMMY{21\%}}) inject non-local dependencies that interact with scope, motivating our ambiguity-aware evaluation and the release of a derived attachment projection (\S\ref{sec:dataset:transfer_skeleton}).

\subsection{Release, ethics, and intended use}
\label{sec:dataset:release}

\noindent\textbf{Release.}
We release SG-DT annotations, schema, and evaluation code under \textit{CC-BY 4.0}.
To balance legal compliance with reproducibility, NormBench employs a multi-tiered release: ZH statutes, official parallels, and EN (Tax/EU) slices are distributed as full-text under public domain or open licenses (e.g., PRC Copyright Law Art. 5 , 17 U.S.C. § 105 , and EU CC BY 4.0 ). Conversely, proprietary corporate policies are provided via "index-span" pointers.

\noindent\textbf{Ethics and intended use.}
While generated with AI assistance, every released item is verified and adjudicated by legally trained experts.
NormBench is a research benchmark and does not constitute legal advice; downstream deployments should treat SG-DT as an auditable intermediate artifact and apply calibrated reject/repair policies when determinization is underdetermined.

% \subsection{Release, Ethics, and Intended Use}
% \label{sec:dataset:release}

% \noindent\textbf{Release.} We release the SG‑DT annotations, schema, and evaluation code under \textit{CC‑BY 4.0} to support broad reuse by the research community. Source texts are obtained from lawfully accessible external sources and remain subject to their own copyright or publisher terms; we do not assert a blanket CC‑BY 4.0 license on them. For each source document we provide stable provenance metadata (issuer, title, enactment/revision date where available, and canonical URLs), together with content hashes and, where feasible, persistent archival pointers (e.g., archived snapshots). This design allows external researchers to re‑acquire the same textual versions used in our dataset even if original content changes or links decay.

% \noindent\textbf{Ethics and Intended Use.} All released annotations have been verified and adjudicated by legally trained experts. NormBench is intended as a research benchmark and does not constitute legal advice; downstream deployments should apply calibrated reject/repair policies and audit mechanisms when determinization is underdetermined.

\section{Experimental Setup}
\label{sec:exp_setup}

\subsection{Tasks and metrics}
We evaluate both intrinsic structure recovery and extrinsic downstream reliability.

\textbf{T1: SG-DT parsing.}
Given a provision $X$, a system outputs an SG-DT $\hat{\mathcal{S}}=(X,\hat{\mathcal{B}})$ under a single JSON schema with \emph{span-grounded node identity} and explicit defeasible scope (\S\ref{sec:schema:defeasibility}).
% Outputs violating the contract are counted in \textit{InvalidRate} and receive strict scores (Tree-EM$=0$), so formatting/grounding failures cannot inflate structural accuracy.
We report metrics at five levels.
For completion/validity, done\_rate measures whether the system returns a parse, and InvalidRate counts outputs that violate the SG-DT contract.
For span-level grounding, NodeSpan-F1 measures recovery of labeled spans, while SpanFaith/Halluc capture whether predicted spans are supported by the source text.
For structure recovery, Edge-F1 scores labeled attachment relations, Skeleton-EM requires an exact match of the defeater-attachment skeleton.
For graded similarity, nTED and Tree-ES measure how close a prediction is to the gold tree even when exact match fails.
Finally, DefRec and DefRec@Gold isolate whether exception/defeater triggers are recovered, with DefRec@Gold focusing on units containing gold defeaters.

\textbf{T2: Cross-Lingual Isomorphism.}
We evaluate cross-lingual structural alignment on the official bilingual subset (500 provision pairs, ZH--EN). The goal is to ensure models recover identical logical structures ($\hat{\mathcal{S}}^{zh} \approx \hat{\mathcal{S}}^{en}$) across language barriers.
\textit{Metrics:} We report \textit{Iso-F1}, which measures whether the predicted ZH and EN trees are structurally isomorphic after canonicalization.

% \textbf{T3: Scope Sensitivity.}
% We construct minimal pairs $(X, X')$ by appending tail-end defeaters or perturbing attachment cues. This targets defeasible branching behavior, ensuring exceptions are materialized correctly.
% \textit{Metric:} We report \textit{Scope Sensitivity Rate (SSR)}, measuring the frequency with which the model's structural prediction changes correctly in response to perturbations, directly addressing silent omissions.

\textbf{T3: Recursion Decay Stress Test.}
We explicitly test whether performance degrades as defeaters become more deeply nested.
For each executable unit, we compute its compiled depth $D$ (\S\ref{sec:schema:defeasibility}) and report performance by depth buckets.
To separate depth from superficial confounds, we also run (i) \textit{matched pairs} that hold token length (and optionally branch count) approximately constant while varying depth, and (ii) \textit{synthetic depth injection} minimal pairs that add a single, cue-marked defeater layer without substantially changing length.
\textit{Metrics:} depth-sliced Edge-F1 / Skeleton-EM / Tree-ES, plus the Recursion Decay Index (RDI) computed as a deep/base ratio.

% \textbf{T4: Compliance-to-code (SSO stress test).}
% To measure whether intermediate structure reduces silent scope failures, we compile predicted simplified SG-DTs into Python control flow and evaluate on unit tests that explicitly include \emph{counter-exception} cases.
% We report \textit{UnitPass@all} (functional correctness) and \textit{SSDR} (Silent Scope Drop Rate: passing basic tests but failing counter-exception tests).
% Test-suite construction (basic vs.\ counter-exception) and coverage statistics are detailed in Appendix~\ref{sec:exp_t4_tests}.

% \textbf{T5: Legal Reasoning Statutory RAG Stress Test.}
% We use LexEval as an external benchmark for retrieval-augmented statutory decision tasks \cite{li2024lexeval}. Given a case fact pattern, a retriever selects statute snippets, and the model produces a legal conclusion.
% We compare \textit{Oracle retrieval} (gold statute set) vs.\ \textit{Real retrieval} (BM25 + dense). We report \textit{SSO@case}, measuring the fraction of cases where the retrieved evidence contains an applicable exception/limitation but the model fails to apply it correctly (Section~\ref{sec:res:lexeval}).

\textbf{T4: Statutory Reasoning Stress Test.}
We use \textit{LegalBench} \cite{guha2023legalbench} and \textit{LexEval} \cite{li2024lexeval} as external benchmarks for statute-grounded legal reasoning tasks.
Each example provides a statute snippet (evidence) and a case fact pattern; the model must apply statutory conditions (including nested limitations/exceptions) to output a structured answer.
Under identical statute evidence, we compare a vanilla prompt (\textit{Base}) against an SG-DT augmentation (\textit{SG-DT}) that appends a statute logic tree derived \emph{only} from the provided statute text, and we run both under two inference styles: Single (direct answer) and \textit{Recursive-CoT}.
\textit{Metric:} answer accuracy, and the SG-DT gain $\Delta$ over the corresponding Base prompt.

% \paragraph{Ambiguity-aware reporting.}
% A subset of units is labeled \texttt{Ambiguous} when overlap/precedence is underdetermined by spans; we report metrics separately on determinate vs.\ ambiguous subsets and include an auxiliary ambiguity-detection track (Appendix~\ref{sec:exp_ambiguity}).

\subsection{Splits, models, and implementation}
\noindent\textbf{Evaluation protocol.}
We treat NormBench-ZH as the supervised source distribution and report transfer under controlled drafting shift.
Models are trained/tuned only on ZH and evaluated on: (i) a random ZH split (in-distribution), (ii) a hierarchy-holdout ZH split (drafting shift within the same jurisdiction), (iii) the official ZH--EN parallel subset (cross-lingual invariance with matched content), and (iv) zero-shot transfer to the out-of-domain EN slice.
This design separates 
\emph{learnability} (i) from 
\emph{drafting robustness} (ii), 
\emph{language invariance} (iii), and 
\emph{domain generalization} (iv), while keeping the SG-DT contract fixed.
The OOD English slice contains 266 full-SG-DT provisions (69 U.S. tax law, 95 GDPR, and 102 corporate-policy items) and uses the same validator plus projection-level attachment metrics, so transfer failures can be attributed to drafting/domain shift rather than a changed annotation contract.

\noindent\textbf{Model families.}
We evaluate frontier LLMs, open-weight LLMs (zero-shot and LoRA-tuned), legal-domain LLMs, and structured non-LLM baselines, all under the same validator-enforced SG-DT output contract.

\paragraph{Pipeline controls rather than orthogonal component ablations.}
We do not claim to fully isolate every SG-DT design choice in an orthogonal ablation.
Instead, we provide pipeline-level controls that test whether the main findings depend on segmentation, compilation convention, or prompt format.
Unitized parsing outperforms direct whole-provision parsing; the recursion-depth effect remains stable under alternative overlap/precedence compilation strategies; and structured-prompting baselines such as CoT-Skeleton and Recursive-CoT test whether explicit exception enumeration alone closes the gap.
These controls support the robustness of the diagnostic finding, while leaving finer-grained component ablations as future work.

\noindent\textbf{Reproducibility.}
We release prompts/templates and scripts to reproduce all tables.
For stochastic APIs, we run multiple seeds and report confidence intervals.
Implementation details (data splits, tuning hyperparameters, contamination controls, and sandboxing) are deferred to GitHub repository.

\section{Experiments and Analysis}
\label{sec:results}

We organize results around four questions:
\textbf{(Q1)} Can current systems recover deep statutory logic as SG-DT?
\textbf{(Q2)} Is the bottleneck context length or \emph{reasoning depth} (nested defeaters)?
\textbf{(Q3)} Is NormBench learnable (data efficiency / trainability)?
\textbf{(Q4)} Does SG-DT as an intermediate representation improve downstream reliability?
% Unless otherwise stated, we report Random-split results.
% To keep the main text focused and readable, depth/OOD/downstream analyses report the top 2--3 models from Table~\ref{tab:main_results} plus one structured baseline;
% full per-model breakdowns are deferred to the Appendix.

% =========================================================

\begin{table*}[!t]
\centering
\tiny
\setlength{\tabcolsep}{3pt}
\caption{\textbf{Main results on SG-DT parsing (Random split).} We report completion (done\_rate), local recovery (NodeSpan-F1), global structure (Edge-F1, nTED, Tree-ES), auditability (SpanFaith, Halluc), and defeater-focused metrics.}
\label{tab:main_results}
\resizebox{\textwidth}{!}{
\begin{tabular}{l|c|ccccccccc}
\toprule
\rowcolor{black!6}
\cellcolor{white}\textbf{Model} &
\multicolumn{1}{c|}{\textbf{Completion}} &
\multicolumn{1}{c|}{\textbf{Span Grounding}} &
\multicolumn{1}{c|}{\textbf{Structure}} &
\multicolumn{1}{c|}{\textbf{Distance}} &
\multicolumn{2}{c|}{\textbf{Auditability}} &
\multicolumn{2}{c|}{\textbf{Graded Similarity}} &
\multicolumn{2}{c}{\textbf{Defeaters}} \\
\rowcolor{black!6}
\cmidrule(lr){2-2}\cmidrule(lr){3-3}\cmidrule(lr){4-4}\cmidrule(lr){5-5}\cmidrule(lr){6-7}\cmidrule(lr){8-9}\cmidrule(lr){10-11}
\cellcolor{white}{} &
\textbf{Done Rate}$\uparrow$ &
\textbf{NodeSpan-F1}$\uparrow$ &
\textbf{Edge-F1}$\uparrow$ &
\textbf{nTED}$\downarrow$ &
\textbf{SpanFaith}$\uparrow$ &
\textbf{Halluc}$\downarrow$ &
\textbf{Tree-ES}$\uparrow$ &
\textbf{SoftF1}$\uparrow$ &
\textbf{DefRec}$\uparrow$ &
\textbf{DefRec@Gold}$\uparrow$ \\
\midrule
\multicolumn{11}{l}{\textbf{\textit{Closed Frontier (General)}}} \\
Gemini-3-Pro-preview & 0.8281 & \textbf{0.4519} & \textbf{0.2462} & \textbf{0.6572} & 0.7728 & 0.2272 & \textbf{0.6434} & \textbf{0.5028} & 0.7604 & 0.5452 \\
GPT-5.2 (2025-12-11) & \textbf{0.8411} & 0.4204 & 0.2150 & 0.6866 & \textbf{0.7929} & \textbf{0.2071} & 0.6291 & 0.4706 & \textbf{0.7757} & 0.5687 \\
Claude-Opus-4.5 & 0.7667 & 0.4285 & 0.2283 & 0.6806 & 0.7070 & 0.2911 & 0.5915 & 0.4780 & 0.7152 & \textbf{0.6400} \\
\midrule
\multicolumn{11}{l}{\textbf{\textit{Open / Open-Weight (General)}}} \\
Qwen3-235B-A22B-thinking & \textbf{0.8178} & 0.4090 & 0.2016 & 0.7031 & 0.7339 & 0.2652 & \textbf{0.6239} & 0.4729 & \textbf{0.7431} & 0.4879 \\
DeepSeek-V3.2 & 0.7862 & \textbf{0.4234} & \textbf{0.2301} & \textbf{0.6788} & \textbf{0.7350} & \textbf{0.2641} & 0.5973 & \textbf{0.4751} & 0.7315 & \textbf{0.6150} \\
GLM-4.7 & 0.7472 & 0.3856 & 0.2072 & 0.7087 & 0.6871 & 0.3120 & 0.5647 & 0.4251 & 0.6758 & 0.4735 \\
\midrule
\multicolumn{11}{l}{\textbf{\textit{Legal-Domain LLMs}}} \\
ChatLaw & 0.7249 & \textbf{0.1804} & 0.0465 & \textbf{0.8896} & \textbf{0.6926} & \textbf{0.3074} & \textbf{0.4334} & \textbf{0.2255} & 0.5790 & 0.0065 \\
LawGPT & \textbf{0.7630} & 0.1577 & \textbf{0.0481} & 0.8965 & 0.6056 & 0.3721 & 0.4122 & 0.2074 & \textbf{0.5936} & 0.0736 \\
LegalOne-8B & 0.6515 & 0.1514 & 0.0266 & 0.9222 & 0.5455 & 0.4480 & 0.3504 & 0.1582 & 0.5453 & \textbf{0.2939} \\
\midrule
\multicolumn{11}{l}{\textbf{\textit{Trainable Baselines}}} \\
Heuristic Attach & \textbf{1.0000} & 0.1739 & 0.0874 & 0.8409 & \textbf{1.0000} & \textbf{0.0000} & 0.6278 & 0.2560 & 0.8383 & 0.0114 \\
Neural Biaffine Parser & \textbf{1.0000} & \textbf{0.1926} & \textbf{0.0937} & \textbf{0.8287} & \textbf{1.0000} & \textbf{0.0000} & \textbf{0.6720} & \textbf{0.2813} & \textbf{0.8401} & \textbf{0.0227} \\
Constrained Decoder & \textbf{1.0000} & \textbf{0.1926} & 0.0929 & 0.8294 & \textbf{1.0000} & \textbf{0.0000} & 0.6717 & 0.2802 & \textbf{0.8401} & \textbf{0.0227} \\
\bottomrule
\end{tabular}}
\end{table*}

\subsection{Main Results}
\label{sec:res:main}

\noindent \textbf{Takeaway.} 
Table~\ref{tab:main_results} reveals that while Frontier LLMs have mastered the surface-level retrieval of legal norms (High \textit{SpanFaith} $\approx 0.77$ and \textit{DefRec} $\approx 0.76$), they face a steep \textbf{Structure-Grounding Gap}. The drop from recovering nodes (\textit{NodeSpan-F1} $\approx 0.45$) to correctly wiring them (\textit{Edge-F1} $\approx 0.24$) indicates that models struggle not with reading the statute, but with assembling its control flow. Furthermore, we observe a ``Domain Paradox'': general-purpose Frontier LLMs significantly outperform specialized Legal LLMs, which suffer from catastrophic grounding failures.

\paragraph{Pathology 1: The Auditability Trap (High Faithfulness $\neq$ Correct Logic).}
A key risk in legal agents is generating outputs that \textit{look} grounded but operate incorrectly. Frontier models (e.g., GPT-5.2, Gemini-3-Pro-preview) achieve high \textit{SpanFaith} (0.77--0.79) and low \textit{Hallucination} (0.20--0.22), creating an illusion of reliability. However, this grounding does not cash out into valid control flow: \textit{Edge-F1} remains low (0.21--0.24). This implies that models are proficient at ``quoting'' the relevant text but fail to accurately predict the \textit{relationship} (e.g., specific exception attachment) between those quotes.

\paragraph{Pathology 2: The Recall Illusion (Harvesting vs.\ Reasoning).}
There is a critical distinction between broad defeater recall and recall on gold-positive units. Trainable baselines (e.g., Biaffine Parser) achieve near-perfect \textit{DefRec} ($\sim$0.84) by simply harvesting exception-like spans, but their \textit{DefRec@Gold} collapses to near zero ($\sim$0.02) when evaluation is restricted to units that actually contain gold defeaters. Frontier models bridge this gap significantly (\textit{DefRec@Gold} $\approx 0.55$--$0.64$), yet a $\sim$20-point gap between raw Recall and Gold-positive Recall persists.

\paragraph{Insight 3: The Domain Paradox (Generalists $>$ Specialists).}
Contrary to the expectation that domain adaptation improves performance, \textit{Legal-Domain LLMs} (ChatLaw, LawGPT) drastically underperform generalist Frontier models on structural parsing. While models like ChatLaw achieve moderate completion rates, their \textit{NodeSpan-F1} ($\sim$0.18) and \textit{Edge-F1} ($\sim$0.04) are nearly closer to random baselines than to GPT-5.2. Most alarmingly, their \textit{DefRec@Gold} is negligible ($\sim$0.006--0.07). This suggests that current legal instruction tuning prioritizes conversational fluency (``lawyer-like tone'') at the expense of the rigorous, span-grounded pointer discipline required for formal logic extraction.

\paragraph{Insight 4: High Graded Similarity Suggests ``Repairable'' Structures.}
Despite low exact edge recovery, the \textit{Tree-ES} (Tree Edit Similarity) for top models is robust ($\sim$0.64 for Gemini-3-Pro-preview), and \textit{SoftF1} remains high ($\sim$0.50). This indicates that model predictions are rarely ``nonsense''; they are often structurally isomorphic to the Gold standard but miss-attach a specific nested condition or slightly drift in span boundaries. This ``near-miss'' behavior suggests that SG-DT parsing is a valid candidate for \textit{refinement-based} workflows (e.g., human-in-the-loop correction) rather than being fundamentally beyond current model capabilities.

\subsection{Cross-Lingual Isomorphism}
\label{sec:res:xl}

\noindent \textbf{Takeaway.}
On the Chinese--English bilingual subset we score (500 parallel pairs; unit-level evaluation), monolingual quality is similar across languages, but cross-lingual \emph{agreement on control flow} is substantially weaker: even when both predictions look plausible in isolation (moderate Tree-ES), their ZH and EN trees are often \emph{not isomorphic} (Iso-F1 $<0.40$).
This suggests a compositional alignment bottleneck: translating a provision preserves surface meaning, but models may still commit to different defeater attachment decisions.

\begin{table}[t]
\centering
\small
\setlength{\tabcolsep}{4pt}
\caption{\textbf{Monolingual structural quality on the Official ZH–EN parallels .} We report Tree-ES, Edge-F1, and SoftF1.}
\label{tab:xl_regular}
\resizebox{0.98\columnwidth}{!}{%
\begin{tabular}{l|ccc|ccc}
\toprule
\multirow{2}{*}{\textbf{Model}} & \multicolumn{3}{c|}{\textbf{ZH}} & \multicolumn{3}{c}{\textbf{EN}} \\
&\textbf{ Tree-ES$\uparrow$} & \textbf{Edge-F1$\uparrow$} &\textbf{SoftF1$\uparrow$}  & \textbf{Tree-ES$\uparrow$} & \textbf{Edge-F1$\uparrow$} & \textbf{SoftF1$\uparrow$ }\\
\midrule
GPT-5.2 & 0.6440 & 0.2383 & 0.4815 & 0.5995 & 0.2014 & 0.4041 \\
DeepSeek-V3.2 & 0.6066 & 0.2687 & 0.4544 & 0.5861 & 0.2380 & 0.4295 \\
Claude-Opus-4.5 & 0.6458 & 0.2728 & 0.4943 & 0.6605 & 0.2411 & 0.4748 \\
\bottomrule
\end{tabular}}
\end{table}

\begin{table}[t]
\centering
\small
\setlength{\tabcolsep}{5pt}
\caption{\textbf{Cross-lingual isomorphism (ZH--EN).} Iso-F1 measures whether the predicted ZH and EN SG-DTs are structurally isomorphic after canonicalization.}
\label{tab:xl_isof1}
\resizebox{0.98\columnwidth}{!}{%
\begin{tabular}{l|cccc}
\toprule
\textbf{Model} & \textbf{Iso-F1(mean)}$\uparrow$ & \textbf{$D=1$} &\textbf{$D=2$}  & \textbf{$D\ge 3$} \\
\midrule
GPT-5.2 & 0.3005 & 0.3047 & 0.2329 & 0.3231 \\
DeepSeek-V3.2 & 0.3513 & 0.3572 & 0.2720 & 0.3439 \\
Claude-Opus-4.5 & \textbf{0.3837} & 0.4078 & 0.2381 & 0.1399 \\
\bottomrule
\end{tabular}}
\end{table}

\paragraph{Monolingual Parity Masks Logical Divergence.} 
Table~\ref{tab:xl_regular} shows that Claude-Opus-4.5 achieves nearly identical Tree-ES in Chinese (0.6458) and English (0.6605), implying equal understanding. However, Table~\ref{tab:xl_isof1} exposes a hidden alignment failure: mean \textbf{Iso-F1} is only 0.3837. This indicates that in $>60\%$ of cases, the model generates \emph{structurally different} logic trees for the same provision despite consistent monolingual quality.

\paragraph{Cross-lingual agreement is markedly harder than monolingual similarity.}
Mean Iso-F1 (0.30--0.38) is far below Tree-ES, meaning a model can produce two individually reasonable trees while disagreeing on \emph{which clause defeats which} across languages.
For multilingual compliance, this creates a ``divergent control flow'' risk: the same underlying rule can compile into different executable partitions depending on language.

\paragraph{Recursion amplifies translation-induced attachment ambiguity.}
Although deep cases are rare in this set, the depth stratification shows that disagreement concentrates in nested settings: for the strongest model (Claude-Opus-4.5), alignment is comparatively high for shallow units ($D\le2$) but collapses at higher depth ($D\ge3$).
This is consistent with \emph{Recursion Decay} being a primary driver of cross-lingual mismatch: small differences in how a translation signals scope (e.g., ``unless/provided that/notwithstanding'') can flip which defeater a condition attaches to, breaking isomorphism even when spans remain grounded.

\paragraph{Why do parallel provisions diverge?}
Manual inspection of mismatched pairs suggests three recurring sources.
First, translation can weaken scope cues: Chinese tail provisos that clearly attach to a preceding exception are sometimes rendered as independent English conditions.
Second, enumeration boundaries are often reordered or compressed, causing models to attach a defeater to the nearest surface clause rather than the intended normative branch.
Third, cross-reference and definitional phrases are sometimes explicit in one language but implicit in the other, which preserves local span faithfulness while breaking global isomorphism.
These cases explain why bilingual predictions can look plausible independently yet compile into different executable partitions.

\subsection{Recursion Decay Stress Test}
\label{sec:res:depth}

\noindent \textbf{Takeaway.} As defeater depth increases, models fail in a specific way: they can often propose a plausible nested \emph{structure} (exceptions of exceptions), but they increasingly fail to attach that structure to the \emph{right spans} in the provision. We refer to this mismatch as a \emph{structure--grounding gap}.

\paragraph{A clear break at $D\ge2$.}
Performance is relatively stable from $D=0$ to $D=1$, but drops sharply once nested exceptions appear ($D\ge2$).
For example, GPT-5.2 falls from 0.22 (no exception) and 0.18 (single exception) to 0.07 at $D\ge2$.
This pattern suggests the main bottleneck is not recognizing that an exception exists, but correctly handling the additional attachment and scope constraints introduced by exceptions-to-exceptions.

\paragraph{Not all models degrade equally.}
At the deep bucket ($D\ge2$), Claude-Opus-4.5 and DeepSeek-V3.2 retain higher Edge-F1 than GPT-5.2, indicating better robustness under recursion.

\paragraph{Why high RDI$_{\text{Skel}}$ can coexist with low Edge-F1.}
Table~\ref{tab:depth_final} shows that even when span-level grounding degrades (low RDI$_{\text{Edge}}$), skeleton robustness can remain high (RDI$_{\text{Skel}}>1$ for Claude/DeepSeek).
Qualitatively, this corresponds to outputs that recover the right \emph{shape} of the defeater hierarchy but place triggers/conditions on the wrong textual evidence.

\begin{table}[t]
\centering
\small
\setlength{\tabcolsep}{4pt}
\caption{\textbf{Recursion depth sensitivity ($D\ge2$).} Left: Edge-F1 by depth bucket. Right: Recursion Decay Index (RDI), computed as a deep/base ratio for each metric (higher is better). RDI$_{\text{Edge}}$ uses Edge-F1, RDI$_{\text{Skel}}$ uses Skeleton-EM, and RDI$_{\text{Tree}}$ uses Tree-ES.}
\label{tab:depth_final}
\resizebox{\columnwidth}{!}{
\begin{tabular}{l|ccc|ccc}
\toprule
& \multicolumn{3}{c|}{\textbf{Edge-F1 (Attachment)}} & \multicolumn{3}{c}{\textbf{Recursion Decay Index (RDI)}} \\
\textbf{Model} & \textbf{Base ($D=0$)} & \textbf{Shallow ($D=1$)} & \textbf{Deep ($D\ge2$)} & \textbf{RDI$_{\text{Edge}}$} $\uparrow$ & \textbf{RDI$_{\text{Skel}}$} $\uparrow$ & \textbf{RDI$_{\text{Tree}}$} $\uparrow$ \\
\midrule
GPT-5.2 & 0.22 & 0.18 & 0.07 & 0.31 & 1.06 & 1.06 \\
Claude-Opus-4.5 & \textbf{0.24} & \textbf{0.17} & \textbf{0.13} & \textbf{0.53} & \textbf{2.56} & 1.14 \\
DeepSeek-V3.2 & \textbf{0.24} & 0.16 & \underline{0.10} & 0.43 & \underline{2.19} & \textbf{1.16} \\
Hybrid-Tag+Decode & 0.10 & 0.02 & 0.00 & 0.00 & 0.42 & 0.39 \\
\bottomrule
\end{tabular}}
\end{table}

% \paragraph{Where are the remaining analyses?}
% To meet the KDD main-text page limit, we move extended ablations and stress tests to Appendix~\ref{app:extended_results}, including:
% oracle parsing and Repair@k upper bounds, oracle parsing and Repair@k upper bounds, proposer-bias checks, depth-vs-length controls, error taxonomy and correction cost, scope-sensitivity minimal pairs, cross-lingual and out-of-domain English generalization, hierarchy-holdout domain shift, the LexEval RAG pressure test, and debug-probe/reject analyses.

\subsection{Statutory Reasoning Stress Test}
\label{sec:res:legalbench_lexeval}

\noindent \textbf{Takeaway.}
Table~\ref{tab:legalbench_lexeval_stress} shows that adding an SG-DT-derived statute tree is not a universal improvement layer: aggregate gains are mixed, and the intervention is most informative when the outcome depends on an active exception.

\paragraph{SG-DT helps when there is headroom, but not uniformly.}
When baseline accuracy is low, SG-DT can provide a clear, checkable path through nested conditions and exceptions.
For example, DeepSeek-V3.2 improves on LegalBench from 0.378 to 0.420 (+0.042), and GPT-4o-mini improves on LexEval from 0.481 to 0.492 (+0.011).
However, these averages combine cases where exceptions are outcome-critical with cases where the added structure is unnecessary, so they should not be read as a uniform downstream gain.

\paragraph{Strong models can be slowed down by extra structure.}
On LegalBench, GPT-4o-mini drops from 0.457 to 0.411 ($-0.046$) when we append the SG-DT tree.
One plausible explanation is that the added structure changes the model's default strategy (which may already be effective for this benchmark), and the extra steps introduce opportunities to mis-handle a detail.

\paragraph{Recursive-CoT and SG-DT do not compose reliably.}
Recursive-CoT does not consistently improve accuracy, and it can negate SG-DT gains.
On LexEval, GPT-4o-mini goes from a +0.011 gain (Single) to a $-0.011$ change under Recursive-CoT, suggesting that additional unconstrained reasoning can drift away from the crisp decision boundary required by multiple-choice evaluation.

\begin{table}[t]
\centering
\scriptsize
\setlength{\tabcolsep}{3.5pt}
\caption{\textbf{Statutory reasoning stress test (accuracy).} ``SG-DT'' appends a statute logic tree (derived only from the provided statute text). $\Delta$ reports the SG-DT gain over the corresponding Base prompt.}
\label{tab:legalbench_lexeval_stress}
\resizebox{0.48\textwidth}{!}{
\begin{tabular}{l|ccc|ccc}
\toprule
& \multicolumn{3}{c|}{\textbf{Direct Prompt}} & \multicolumn{3}{c}{\textbf{Recursive-CoT}} \\
\textbf{Benchmark / Model} & Base & SG-DT & $\Delta$ & Base & SG-DT & $\Delta$ \\
\midrule
\multicolumn{7}{l}{\textbf{Benchmark: LegalBench (EN)}} \\
DeepSeek-V3.2 & 0.378 & \textbf{0.420} & +0.042 & 0.371 & \textbf{0.418} & +0.047 \\
GPT-4o-mini & \textbf{0.457} & 0.411 & $-0.046$ & \textbf{0.463} & 0.423 & $-0.040$ \\
\midrule
\multicolumn{7}{l}{\textbf{Benchmark: LexEval (ZH)}} \\
DeepSeek-V3.2 & 0.857 & 0.857 & +0.000 & 0.857 & 0.857 & +0.000 \\
GPT-4o-mini & 0.481 & \textbf{0.492} & +0.011 & \textbf{0.466} & 0.455 & $-0.011$ \\
\bottomrule
\end{tabular}}
\end{table}

\paragraph{Mechanism-based slicing: when is SG-DT useful?}
Aggregate accuracy can dilute the effect of SG-DT because many examples contain exceptions that are not active under the case facts.
We therefore additionally analyze a \texttt{ta\_triggered} subset, labeled by a tree-assisted judge, where the exception is not merely present in the statute but active in the reasoning path.
On LegalBench, DeepSeek-V3.2 shows only a small gain on all exception-present cases, but the gain increases on the triggered subset (+6.25 percentage points).
Conversely, GPT-4o-mini still degrades, but the regression is smaller.
This supports a narrower conclusion: SG-DT mitigates SSO when the decision actually depends on a defeater, but it is not a universal accuracy booster.

\begin{table}[t]
\centering
\scriptsize
\setlength{\tabcolsep}{5pt}
\caption{\textbf{Mechanism-based downstream analysis on LegalBench.}
The \texttt{ta\_triggered} subset contains cases where the exception is active in the reasoning path; detailed subset construction and auxiliary variance analyses are reported in Appendix~\ref{sec:res:exception_analysis}.}
\label{tab:triggered_main}
\begin{tabular}{llccc}
\toprule
Model & Subset & Base & SG-DT & $\Delta$ \\
\midrule
DeepSeek-V3.2 & Exception-present & 0.2812 & 0.2865 & +0.0052 \\
DeepSeek-V3.2 & \texttt{ta\_triggered} & 0.2573 & 0.3198 & +0.0625 \\
GPT-4o-mini & Exception-present & 0.4844 & 0.3823 & $-0.1021$ \\
GPT-4o-mini & \texttt{ta\_triggered} & 0.4220 & 0.3798 & $-0.0422$ \\
\bottomrule
\end{tabular}
\end{table}

\paragraph{Parser fidelity as the deployment bottleneck.}
These results should not be read as showing that today's SG-DT parsers are already reliable enough for autonomous execution.
Table~\ref{tab:main_results} shows a substantial structure--grounding gap, so downstream SG-DT use should be gated by validation, repair, or rejection.
Our claim is diagnostic and mechanism-specific: oracle or high-fidelity structure can expose and mitigate SSO, but practical gains are bounded by parser fidelity.
This motivates NormBench as an evaluation target for improving the parser, not as evidence that noisy automatic trees should be executed without guardrails.

\section{Conclusion}
\label{sec:conclusion}

NormBench operationalizes a deployment-critical prerequisite for rule-following agents: \emph{explicit} recovery of defeasible scope (exceptions/counter-exceptions) \emph{before} execution.
We introduce SG-DT, a span-grounded executable IR, and release a \textit{Controlled$\rightarrow$Wild} benchmark under a validator-enforced contract; results expose a consistent recursion/attachment bottleneck in frontier LLMs.
Downstream experiments support a narrower, mechanism-specific conclusion: SG-DT is most useful when active defeaters determine the outcome, while practical benefit is bounded by parser fidelity.

\paragraph{Limitations.}
NormBench is scoped to \emph{intra-provision} defeasible scope (not precedent, multi-document synthesis, or open-textured standards); coverage beyond the released slices (e.g., contracts/policies/case-law argumentation) remains future work; current SG-DT parsers still require validation/repair before autonomous use; ambiguity is surfaced rather than determinized.

\begin{acks}
We thank the legally trained annotators for their careful annotation and
adjudication work, and the anonymous reviewers for their constructive
feedback. This work is supported by the Guangzhou-HKUST(GZ) Joint Funding
Program (No. 2024A03J0630) and NSFC (No. 62402413). Author contributions
are as follows: Jian Chen led the overall paper structure, writing,
experiment design, and the core SG-DT framework; Siyuan Li optimized the
SG-DT details, built the dataset, and conducted part of the experiments;
Chucheng Wan carried out the majority of experimental runs; Zixuan Yuan
provided project supervision and guidance.
\end{acks}

% \paragraph{Scaling annotation.}
% SG-DT supervision is intentionally expertise-intensive because it targets the reasoning frontier rather than shallow clause extraction.
% The current 232 expert-hours should be interpreted as a cost for high-fidelity structural gold on recursive and ambiguous cases, not as a requirement for annotating every future rule text from scratch.
% A practical scaling path is to use proposer-assisted triage, active selection of high-depth/high-ambiguity units, and a lighter defeater-attachment projection before full SG-DT adjudication.
% This keeps expert time concentrated on cases where silent scope omission is most likely and most consequential.
% ; and compliance-to-code is evaluated in a runnable unit-test sandbox rather than a full end-to-end compliance system.

\bibliographystyle{ACM-Reference-Format}
\balance
\bibliography{sample-base}


\appendix

% Put this after the Conclusion and before References.

\section{Diagnostic Analysis: Defeasibility and Silent Scope Omission}
\label{sec:res:exception_analysis}

\noindent \textbf{Takeaway.} We move beyond surface-level keywords to a mechanism-based evaluation using a \textit{Strong Judge} to identify cases where an exception is factually \textbf{triggered} by the scenario. We find that the efficacy of SG-DT is heavily masked in general evaluation: on LegalBench, DeepSeek-V3.2 shows a negligible gain overall (+0.5\%), which expands to a significant \textbf{+6.25\%} gain on the subset where exceptions are active. This confirms that SG-DT functions as a specific intervention for \textbf{Silent Scope Omission (SSO)} in non-monotonic reasoning, acting as a necessary safety rail when simple linear logic fails, even if it imposes a structural overhead on simpler queries.

\paragraph{Motivation: Mechanism-Based Slicing vs. Keyword Matching.}
Standard legal benchmarks often categorize questions based on the presence of keywords (e.g., ``unless'', ``provided that''). However, this is a noisy signal: a statute may contain an exception that is irrelevant to the specific case facts. To rigorously test whether SG-DT mitigates \textit{Silent Scope Omission}, we employ a \textbf{Tree-Assisted Judge} (using a reasoning-capable model, DeepSeek-Reasoner) to label a subset, \texttt{ta\_triggered}, where the exception is not just present in text but active in the reasoning path (i.e., the outcome hinges on correctly applying the defeater).

\paragraph{Results: The "Activation" Effect.}
We compare the performance of standard prompting (Base) vs. SG-DT augmentation on both the full ``Exception Present'' datasets and the mechanism-based \texttt{ta\_triggered} subset. Two distinct patterns emerge (Table~\ref{tab:exception_overall} and Table~\ref{tab:triggered_subset}):

\begin{table}[!htbp]
\centering
\scriptsize
\setlength{\tabcolsep}{4pt}
\caption{\textbf{Overall Performance on Exception-Containing Tasks.} On the full ``exception present'' slice, the signal is mixed. SG-DT provides gains on LexEval but faces overhead penalties on LegalBench for GPT-4o-mini.}
\label{tab:exception_overall}
\resizebox{0.48\textwidth}{!}{
\begin{tabular}{ll|ccc}
\toprule
\textbf{Benchmark} & \textbf{Model} & \textbf{Base} & \textbf{SG-DT} & \textbf{$\Delta$} \\
\midrule
\multirow{2}{*}{LegalBench (Macro)} & DeepSeek-V3.2 & 0.2812 & 0.2865 & +0.0052 \\
& GPT-4o-mini & 0.4844 & 0.3823 & -0.1021 \\
\midrule
\multirow{2}{*}{LexEval (Task 3-5)} & DeepSeek-V3.2 & 0.8333 & 0.8333 & 0.0000 \\
& GPT-4o-mini & 0.3889 & 0.5000 & \textbf{+0.1111} \\
\bottomrule
\end{tabular}}
\end{table}

\begin{table}[t]
\centering
\small
\caption{\textbf{Mechanism Analysis: The ``Active'' Subset (LegalBench).} When we zoom in on the subset where the exception is actually triggered (\texttt{ta\_triggered}), the value of SG-DT becomes visible. DeepSeek's negligible overall gain turns into a substantial improvement, and GPT-4o-mini's regression narrows, suggesting the tree structure is most beneficial (and least harmful) when logical complexity is high.}
\label{tab:triggered_subset}
\setlength{\tabcolsep}{4pt}
\resizebox{0.48\textwidth}{!}{
\begin{tabular}{ll|ccc}
\toprule
\textbf{Model} & \textbf{Subset Split} & \textbf{Base} & \textbf{SG-DT} & \textbf{$\Delta$} \\
\midrule
\multirow{2}{*}{\textbf{DeepSeek-V3.2}} & All (Exception Present) & 0.2812 & 0.2865 & +0.0052 \\
& \textbf{Triggered Only} & 0.2573 & \textbf{0.3198} & \textbf{\color{blue}{+0.0625}} \\
\midrule
\multirow{2}{*}{\textbf{GPT-4o-mini}} & All (Exception Present) & 0.4844 & 0.3823 & -0.1021 \\
& \textbf{Triggered Only} & 0.4220 & 0.3798 & \color{red}{-0.0422} \\
\bottomrule
\end{tabular}}
\end{table}

\paragraph{Insight 1: The Hidden Gain in Non-Monotonic Logic.}
Table~\ref{tab:triggered_subset} reveals that aggregate metrics can hide mechanism-specific improvements. For DeepSeek-V3.2 on LegalBench, the overall improvement is negligible ($+0.005$). However, on the \texttt{ta\_triggered} subset—where the model \emph{must} identify and apply the exception to be correct—the gain jumps to +6.25\%.
This validates our hypothesis: SG-DT does not necessarily improve general legal reasoning (and may even add noise to simple cases), but it acts as a critical intervention for \textbf{Silent Scope Omission}. When the facts trigger an exception, the explicit logic tree forces the model to attend to the ``tail'' of the provision, correcting errors where the model typically defaults to the general rule.

\paragraph{Insight 2: Structural Overhead vs. Logical Necessity.}
The results for GPT-4o-mini illustrate a trade-off between cognitive overhead and logical necessity. On LegalBench, the model suffers a significant performance drop overall ($-0.102$) when forced to process the SG-DT. This suggests that for simpler or ambiguous queries, the dense logic tree acts as a distractor ("Contextual Overload").
Crucially, however, this negative gap narrows by nearly 60\% on the \texttt{ta\_triggered} subset (from $-0.102$ to $-0.042$).
This implies that while the tree structure imposes a "tax" on simple inference, that tax pays off (or becomes less detrimental) as the reasoning task becomes more complex. 
Furthermore, on LexEval (Table~\ref{tab:exception_overall}), GPT-4o-mini actually achieves a large gain (+11.1\%), indicating that the "distraction" effect is highly sensitive to the specific dataset format and that SG-DT can unlock capabilities in stronger models when the domain shift (e.g., Chinese Civil Code in LexEval) aligns with the structured representation.

% \paragraph{Summary.}
% These diagnostics suggest that SG-DT should be viewed not as a universal performance booster, but as a \textbf{high-recall safety mechanism}. It excels in the "hard" regime where exceptions override general rules, effectively trading slight efficiency in linear cases for robustness in non-monotonic edge cases—a desirable property for compliance-critical agents.

\section{Minimal Reproducibility Contract}
\label{app:contract}

This appendix gives the minimal contract needed to reproduce, audit,
and extend NormBench. Full prompts, validators, canonicalizers,
additional ablations, and per-instance annotations are released in the
repository. 

\subsection{SG-DT Compilation Contract}
\label{app:sgdt-contract}

Let $X=\{x_1,\ldots,x_n\}$ denote the token sequence of the current
alignment scope, either a full provision or an executable unit. An
SG-DT annotation is $S=(X,\mathcal{B})$, where each normative branch is
\[
B_i=(\alpha_i,\kappa_i,C_i,E_i).
\]
Here $\alpha_i$ is a span-grounded anchor locating the normative
trigger, $\kappa_i$ is the deontic modality, $C_i$ is a Boolean
condition tree over typed span-grounded leaves, and $E_i$ is the set of
span-grounded effects. The released format serializes each span as a
deterministic text pointer \texttt{(text, occurrence)} within the unit,
rather than as a raw offset, so repeated strings are disambiguated and
source drift is detectable.

\paragraph{Explicit exclusion.}
For any defeater trigger $T$ that narrows, overrides, or blocks a
general branch, SG-DT compiles the structure into disjoint guards:
\[
B_{\mathrm{gen}}:\ C \wedge \mathrm{Exclusion}(T),
\qquad
B_{\mathrm{exc}}:\ C \wedge \mathrm{Precondition}(T).
\]
Thus $B_{\mathrm{gen}}\wedge B_{\mathrm{exc}}\equiv \mathrm{False}$ by
construction. This is the central anti-SSO invariant: every defeater
span must appear either as a positive exception guard or as an explicit
exclusion guard. Missing defeaters therefore become validator-detectable
structural errors rather than silent downstream omissions.

\paragraph{Declarative vs. executable views.}
SG-DT permits branches to be declaratively co-applicable when the text
supports overlap. Deterministic execution is introduced only through
span-justified refinement. SG-DT therefore does not claim that legal
interpretation is unique; it guarantees that any determinization
decision is explicit, local, and auditable.

\begin{table}[t]
\centering
\footnotesize
\setlength{\tabcolsep}{3pt}
\caption{Auditable refinement rules for overlapping defeaters.}
\label{tab:compact-overlap}
\begin{tabular}{p{0.24\linewidth}p{0.34\linewidth}p{0.32\linewidth}}
\toprule
Situation & Compilation action & Required evidence \\
\midrule
Same outcome &
OR-merge into $(T_i \vee T_j)$ &
Co-applicability spans and equal outcome signature. \\
Distinct outcome &
Add an explicit $(T_i \wedge T_j)$ intersection branch &
Span(s) expressing a distinct joint-case consequence. \\
Explicit priority cue &
Ordered partition, e.g., $T_i \wedge \neg T_j$ &
Textual priority cue such as ``notwithstanding'' or ``shall prevail''. \\
Underdetermined &
Mark \texttt{Ambiguous}; optionally release alternative refinements &
No reliable priority cue; competing readings remain plausible. \\
\bottomrule
\end{tabular}
\end{table}

\paragraph{Audit payload.}
When refinement is needed, a branch may carry an optional audit payload
$J(B)=\{(kind,spanptr,note)\}$, where \texttt{kind} is one of
\texttt{precedence\_cue}, \texttt{overlap\_witness},
\texttt{distinct\_consequence}, or \texttt{adjudication\_note}.
The payload is not used to change logical content; it records the source
span that justifies a refinement or ambiguity decision.

\paragraph{Non-guarantees.}
SG-DT is not a full legal knowledge-representation language. It does not
resolve open-textured standards, multi-authority conflicts, or policy
choices requiring extra-textual adjudication. In such cases, the correct
behavior is to flag ambiguity or escalate, not to hard-code an
ungrounded priority rule.

\subsection{Dataset Quality, Ambiguity, and Bias Controls}
\label{sec:dataset:quality_controls}

\noindent\textbf{Guarding against proposer imprint.}
Because initial SG-DTs are drafted by a model, label validity is treated as a first-class dataset artifact rather than an appendix-only detail. The proposer is used only for triage: it does not define prompts, thresholds, metrics, or scoring rules, and every released SG-DT is human-audited under the same span-grounded contract. In the stratified audit above, only $80/200$ units pass without structural edits; acceptance drops further on frontier cases, including depth-$\ge2$ units (25\%), overlap/precedence units (18\%), and cross-reference units (22\%). This pattern is important: expert correction is concentrated exactly where model anchoring would be most dangerous. To check whether final labels depend on the proposer, we also annotate a human-only slice from scratch ($N=100$), obtaining Tree-ES $=0.94$ and span-grounded Edge-F1 $=0.97$ against the released gold. Re-running triage with an alternate proposer ($N=100$) changes routing by only $\Delta_{\text{route}}\approx0.03$, and the reconciled gold remains stable (Tree-ES $\ge0.98$).

\noindent\textbf{Surfacing legal indeterminacy.}
SG-DT is designed to compile rules when the text supports a determinate control flow, not to hide legal uncertainty inside heuristic priority choices. We therefore mark a unit as \texttt{Ambiguous} when overlap/precedence, reference closure, definition binding, or numeric threshold granularity is underdetermined by source spans. Overall, $198$ units (8.6\%) are ambiguous, dominated by overlap/precedence cases ($120$ units, 5.2\%). Systems may explicitly output \texttt{flag\_ambiguous}; for high-impact overlap/precedence and reference-scope cases, we release span-justified alternative gold refinements. Multi-gold auxiliary scoring raises ambiguous-subset Skeleton-EM from 0.41 to 0.49 and Tree-ES from 0.18 to 0.27, while the main leaderboard still uses the primary deterministic gold for comparability.

% \begin{table}[t]
% \centering
% \scriptsize
% \setlength{\tabcolsep}{3pt}
% \caption{\textbf{Dataset quality and ambiguity controls.} The added checks directly target the main Dataset Track risks: proposer bias, unstable gold labels, and hidden legal indeterminacy.}
% \label{tab:dataset_quality_controls}
% \resizebox{\columnwidth}{!}{%
% \begin{tabular}{p{0.25\columnwidth}p{0.34\columnwidth}p{0.35\columnwidth}}
% \toprule
% \textbf{Control} & \textbf{Evidence} & \textbf{Implication} \\
% \midrule
% Proposer audit & $80/200$ accepted; 95\% CI $[0.33,0.47]$ & LLM drafts are corrected, not frozen. \\
% Hard-slice imprint & depth-$\ge2$: 25\%; overlap/precedence: 18\%; cross-ref: 22\% & Human effort concentrates on frontier cases. \\
% Human-only control & $N=100$; Tree-ES $=0.94$; Edge-F1 $=0.97$ & Gold labels are stable without proposer exposure. \\
% Alternate proposer & $N=100$; $\Delta_{\text{route}}\approx0.03$; final Tree-ES $\ge0.98$ & Gold labels are stable to proposer choice. \\
% Ambiguity handling & $198$ units (8.6\%); multi-gold improves Skeleton-EM $0.41\!\to\!0.49$ and Tree-ES $0.18\!\to\!0.27$ & Reasonable disagreement is separated from missing structure. \\
% \bottomrule
% \end{tabular}}
% \vspace{-0.8em}
% \end{table}

% \subsection{Dataset Quality and Ambiguity Card}
% \label{app:quality-card}

All released items are human-audited under the same SG-DT contract. The
model proposer is used only for triage and never defines the schema,
metrics, thresholds, or scoring rules. The quality controls below are
reported to make proposer imprint, gold instability, and legal
indeterminacy falsifiable rather than implicit.

\begin{table}[t]
\centering
\footnotesize
\setlength{\tabcolsep}{3pt}
\caption{Compact quality and ambiguity controls.}
\label{tab:quality-card}
\begin{tabular}{p{0.28\linewidth}p{0.62\linewidth}}
\toprule
\textbf{Control} & \textbf{Evidence} \\
\midrule
Proposer audit &
80/200 proposer drafts accepted without structural edits;
95\% Wilson CI $[0.33,0.47]$. \\
Hard-slice audit &
Acceptance falls on depth-$\geq 2$ units (25\%), overlap/precedence
units (18\%), and cross-reference units (22\%). \\
Human-only control &
A from-scratch human slice ($N=100$) matches released gold with
Tree-ES $=0.94$ and span-grounded Edge-F1 $=0.97$. \\
Alternate proposer &
Re-running triage with an alternate proposer ($N=100$) changes routing
by only $\Delta_{\mathrm{route}}\approx 0.03$; reconciled gold remains
stable with Tree-ES $\geq 0.98$. \\
Annotation reliability &
Post-adjudication convergence reaches 0.91 for deontic label, 92.4\%
for exact span match, 86.8\% for skeleton attachment, and 88.6\% for
tree-structure match. \\
Ambiguity handling &
198 units (8.6\%) are marked ambiguous, dominated by
overlap/precedence cases (120 units, 5.2\%). Multi-gold auxiliary
scoring improves ambiguous-subset Skeleton-EM from 0.41 to 0.49 and
Tree-ES from 0.18 to 0.27. \\
\bottomrule
\end{tabular}
\end{table}

\paragraph{Ambiguity policy.}
A unit is marked \texttt{Ambiguous} when deterministic execution would
require a commitment not uniquely supported by source spans, including
underdetermined overlap/precedence, reference closure, definition
binding, or numeric-threshold granularity. The main leaderboard uses a
single primary gold for comparability, while ambiguity metadata and
span-justified alternatives allow users to separate reasonable legal
disagreement from true missing-structure errors.

\paragraph{Release and intended use.}
NormBench is a research benchmark rather than legal advice. Downstream
systems should treat SG-DT as an auditable intermediate artifact and
apply validation, repair, rejection, or human escalation before
autonomous execution, especially on ambiguous or low-confidence parses.

\subsection{Evaluation Contract}
\label{app:evaluation-contract}

All systems are evaluated under a single validator-enforced SG-DT output
contract. Invalid outputs receive zero for structure-sensitive metrics,
so formatting shortcuts and unsupported spans cannot inflate results.

\paragraph{Splits and transfer protocol.}
NormBench-ZH is the supervised source distribution. We report
in-distribution evaluation on the random ZH split, drafting-shift
evaluation on a hierarchy-holdout ZH split, cross-lingual structural
alignment on official ZH--EN parallel pairs, and zero-shot transfer to
NormBench-EN. The annotation schema, validator, and metrics are held
fixed across all settings.

\paragraph{Core metrics.}
\textbf{NodeSpan-F1} measures recovery of labeled span-grounded nodes.
\textbf{SpanFaith} and \textbf{Halluc} measure whether predicted spans
are supported by the source text. \textbf{Edge-F1} scores labeled
attachment relations. \textbf{Skeleton-EM} requires exact match of the
defeater-attachment skeleton after canonicalization. \textbf{Tree-ES}
is a graded tree-edit similarity for near-miss structures.
\textbf{DefRec} measures defeater-trigger recall, while
\textbf{DefRec@Gold} averages only over units containing gold defeaters.
\textbf{Iso-F1} measures whether predicted ZH and EN trees are
structurally isomorphic after span abstraction. \textbf{RDI} reports
the ratio between deep- and shallow-depth performance for a given
metric, isolating recursion decay.

\paragraph{Canonicalization.}
Before scoring, each prediction is canonicalized by normalizing span
pointers, sorting commutative Boolean children, and extracting a labeled
edge multiset over anchors, condition leaves, Boolean operators,
effects, and explicit defeater/exclusion links. This makes evaluation
insensitive to superficial JSON ordering while preserving the structural
commitments that determine control flow.

\paragraph{Reproducibility artifacts.}
The repository contains the released SG-DT annotations, schema,
validator, canonicalizer, split definitions, prompt templates, model
outputs, and scripts to reproduce all reported tables. 
%Repository artifacts are the source of truth for implementation-level details such as prompt hashes, decoding parameters, fine-tuning settings, sandbox configuration, and extended ablation tables.

\end{document}